%% file: root.tex
\documentclass[letterpaper, 10 pt, conference]{ieeeconf}  % Comment this line out if you need a4paper

\IEEEoverridecommandlockouts                              % This command is only needed if 
\usepackage{graphics} % for pdf, bitmapped graphics files
\usepackage{epsfig} % for postscript graphics files
\usepackage[utf8]{inputenc} % allow utf-8 input
\usepackage[T1]{fontenc}    % use 8-bit T1 fonts
\usepackage{url}            % simple URL typesetting
\usepackage{booktabs}       % professional-quality tables
\usepackage{amsfonts}       % blackboard math symbols
\usepackage{nicefrac}       % compact symbols for 1/2, etc.
\usepackage{microtype}      % microtypography
\usepackage{graphicx}
\usepackage{grffile}
\usepackage{amsmath, amssymb}
\usepackage{algpseudocode}
\usepackage{subcaption}
\usepackage{multirow}
\usepackage{multicol}
\usepackage{bm}
\usepackage{color}

\usepackage{epstopdf}

\usepackage[table]{xcolor}
\usepackage{float}
\usepackage{array}
\usepackage{comment}

\makeatletter
\let\NAT@parse\undefined
\makeatother
\usepackage{hyperref}
\hypersetup{
    colorlinks=true,
    urlcolor=blue,
}
\makeatletter
\def\footnoterule{\kern-3\p@
  \hrule \@width 2in \kern 2.6\p@} % the \hrule is .4pt high
\makeatother

\title{\LARGE \bf
Object Memory Transformer for Object Goal Navigation
}

\makeatletter
\let\@oldmaketitle\@maketitle% Store \@maketitle
\renewcommand{\@maketitle}{\@oldmaketitle% Update \@maketitle to insert..

\centering
  \includegraphics[width=0.85\textwidth]{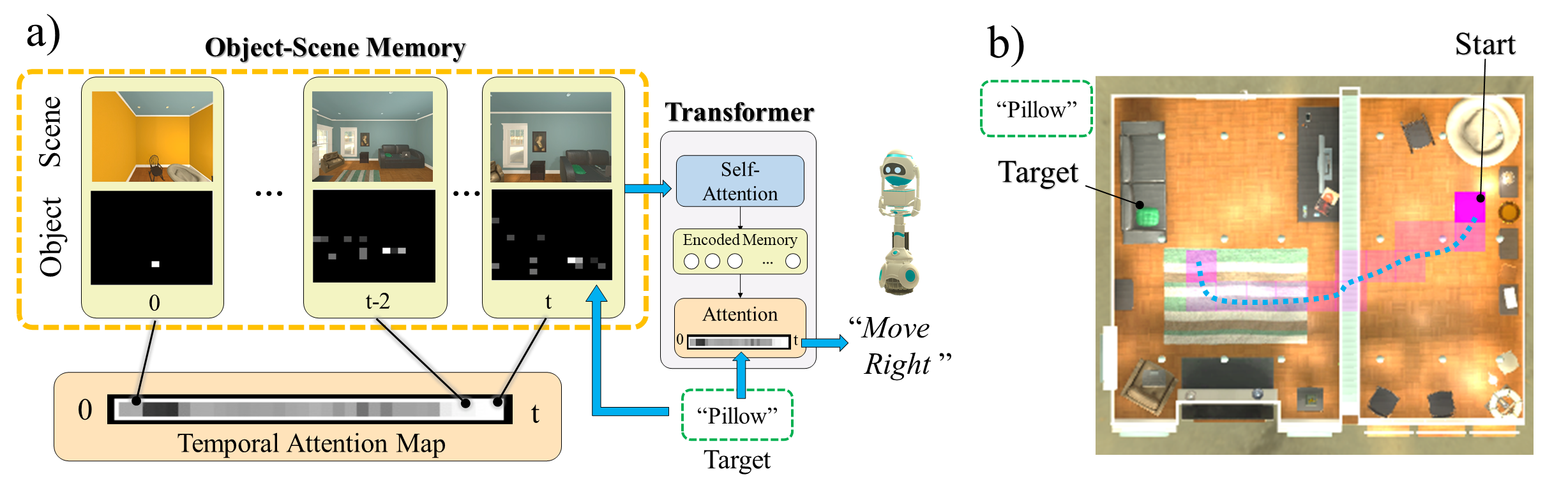}
  \captionof{figure}{a) Object Memory Transformer (OMT) for efficient indoor navigation to find the target object. Here, the agent looks for ``Pillow'' on the sofa, which is in the next room and the agent cannot observe it from the start position. Even in this complex case, OMT can take an efficient path to the target shown as a trajectory in the blue dots in (b) by exploiting long-term visual cues while taking into account the relevance of the feature at each time step. Specifically, OMT stores long-term history of observed scenes and objects into Object-Scene Memory (OSM), attends to the crucial ones in OSM, and produces useful features for the RL agent. At the begging of the episode, when the agent is looking at a wall, it rarely attends to those histories (notice at the dark gray region in the attention map of (a)). Once the target comes into the agent's sight, the attention weight gets higher (shown in white).}
  \label{figurelabel}
  }
 \vspace{-7mm}

\makeatother

\author{Rui Fukushima$^{1}$, Kei Ota$^{2,3}$, Asako Kanezaki$^{2}$, Yoko Sasaki$^{1}$ and Yusuke Yoshiyasu$^{1}$% <-this % stops a space
\thanks{$^{1}$Rui Fukushima, Yoko Sasaki and Yusuke Yoshiyasu are with National Institute of Advanced Industrial Science and Technology (AIST), Japan
        {\tt\small rui.fukushima, y-sasaki, yusuke-yoshiyasu@aist.go.jp}}%
\thanks{$^{2}$Kei Ota and Asako Kanezaki are with Tokyo Institute of Technology, Tokyo, Japan
        {\tt\small kanezaki@c.titech.ac.jp}}%
\thanks{$^{3}$Kei Ota is with Information Technology R\&D Center, Mitsubishi Electric Corporation, Kanagawa, Japan.
        {\tt\small Ota.Kei@ds.MitsubishiElectric.co.jp}}
}

\begin{document}
\maketitle
\thispagestyle{empty}
\pagestyle{empty}
\input{abstract.tex}
\input{introduction}
\input{related_works}

\input{proposed_method}
\input{experiments}

\input{results_discussion}

\input{conclusion}
\section*{Acknowledgment}
This paper is in part based on the results obtained from a project, JPNP20006, commissioned by the New Energy and Industrial Technology Development Organization (NEDO).

%The preferred spelling of the word ÒacknowledgmentÓ in America is without an ÒeÓ after the ÒgÓ. Avoid the stilted expression, ÒOne of us (R. B. G.) thanks . . .Ó  Instead, try ÒR. B. G. thanksÓ. Put sponsor acknowledgments in the unnumbered footnote on the first page.

%References are important to the reader; therefore, each citation must be complete and correct. If at all possible, references should be commonly available publications.
%%%%%%%%%%%%%%%%%%%%%%%%%%%%%%%%%%%%%%%%%%%%%%%%%%%%%%%%%%%%%%%%%%%%%%%%%%%%%%%%
\bibliographystyle{IEEEtran}
\bibliography{reference}

%\rui{
\section*{Appendix}
%Appendixes should appear before the acknowledgment.
\subsection{Implementation Details}
Our framework is implemented by using PyTorch, as well as Transformer was implemented according to provided PyTorch package. RMSprop was used for the optimization. The learning rate was $7\times10^{-4}$, linearly decreasing to $0$ in the last epoch. We use one layer in the encoder and decoder, and five multi heads attention in Transformer. The model was trained with $80$ threads with $25$ million frames in total with four NVIDIA V100 GPUs. For the evaluation the agent is randomly located at the beginning of each episode and expected to reach the target object within $300$ steps. we terminate the episode if the total number of steps exceeds it, and consider it to be failure. We evaluate our model with $250$ episodes on each target from target class sets $G$.
%}

\end{document}

%% file: abstract.tex
\begin{abstract}
This paper presents a reinforcement learning method for object goal navigation (ObjNav) where an agent navigates in 3D indoor environments to reach a target object based on long-term observations of objects and scenes. 
% To this end, we propose Object Memory Transformer (OMT) that attends to salient objects in the sequence of observed scenes and objects that are stored in a memory called Object-Scene Memory. 
To this end, we propose Object Memory Transformer (OMT) that consists of two key ideas: 1) Object-Scene Memory (OSM) that enables to store long-term scenes and object semantics, and 2) Transformer that attends to salient objects in the sequence of previously observed scenes and objects stored in OSM.
%This mechanism can be thought of as landmarks to guide the agent and 
This mechanism allows the agent to efficiently navigate in the indoor environment without prior knowledge about the environments, such as topological maps or 3D meshes.
To the best of our knowledge, this is the first work that uses a long-term memory of object semantics in a goal-oriented navigation task. Experimental results conducted on the AI2-THOR dataset show that OMT outperforms previous approaches in navigating in unknown environments. In particular, we show that utilizing the long-term object semantics information improves the efficiency of navigation.

%It is widely known that the use of landmarks detected in the environment for navigation tasks is very effective in accomplishing the task. For instance, in mammalian navigation, the information corresponding to "What", represented by landmarks \textit{etc.}, is as important as the spatiotemporal information corresponding to "Where" and "When". On the other hand, it is not yet known the effect of inputting the long-term these objects information. The purpose of this study is to investigate the effects of long-term spatiotemporal and landmark information interaction on robot navigation.
%The purpose of this research is to propose a visual navigation framework that utilize  spaciotemporal and landmark information. 
%We achieve this goal by encoding spatiotemporal and landmark information into an external memory named "Object-Scene Memory" that can store long-term temporal information. To the best of our knowledge, this is the first time that not only observations but also long-term objects information has been implemented in a target-driven navigation task without  Recurrent Neural Networks (RNNs). Our experiments on AI2-THOR shows the proposed model achieves higher performance and effectiveness in new scene compared to previous approaches. In particular, long-term object information was indicated to be involved in improving the effectiveness of navigation. At the end of this paper, a qualitative comparison with the atate-of-the-art (SOTA) is presented.

\end{abstract}

%% file: introduction.tex
\section{Introduction}
Object goal navigation (ObjNav) is a task of navigating an agent toward a target object given by a word with reference to the first-person view \cite{Wei2018,Druon2020}. To solve the task, a navigation agent is required to learn highly nonlinear mappings from images and words to actions. Unlike in game-playing tasks such as GO, where Deep Reinforcement Learning (DRL) agents defeated human professionals~\cite{Silver2016,Badia2020}, DRL agents' performances on ObjNav are still far behind from that of average humans~\cite{Druon2020}.

%Deep Neural Networks (DNNs) have achieved superior results over humans in various games such as GO and Atari \cite{Silver2016,Badia2020}. However, spatial navigation with Deep Reinforcement Learning (DRL) is far behind that of mammals in several aspects such as learning efficiency and generalizability \cite{Bermudez-Contreras2020}. The mechanisms that allow mammals to achieve such a high capacity for spacial recognition are not yet fully understood, but some studies have suggested that landmarks such as objects are involved \cite{Meilinger2016,Morris1981,janneke2015,McNaughton1991,Malcolm2018,Oyvind2019}. Recently, robot navigation using object information has been actively worked on, and also there has been remarkable improvement in performance \cite{Druon2020,Wei2018,Qiu2020,Du2021,Lv2020,Mayo2021}.

The main challenge in ObjNav is that the task requires complex scene understanding and can sometimes become long-horizon. In many cases, the target object is not visible from the start position and thus the agent needs to get close to the object by understanding the surrounding environment to plan its paths from the first-person views in which the goal may not be present. Further, these views drastically change according to the agent’s moves and sometimes do not contain any meaningful information. For example, when the agent faces to a white wall, %the agent's behavior can be unstable since 
there is no visual clue in the view to decide the next move. % in the next steps.% \rui{where no clue of which way to go}, which makes the agent’s behavior unstable. 
As a result, the learning agent may easily be stuck and pivot at the same place almost forever or may take an unexpectedly none-smooth trajectory that repeatedly arrives at the same locations back and forth. Another difficult point in ObjNav is its diversity: the trained agent model is required to generalize to unknown environments with different room structures, scenes, objects, etc., which makes this task even harder.
% Note that in ObjNav an agent is not only required to travel in a known environment but also needs to generalize to unknown environments, which makes this task even harder.

%However, the object information used in these studies is independent from the temporal context. Therefore, information about the objects seen in the episode or the order in which the objects appeared are missing. Although some of these studies have used Recurrent Neural Networks (RNNs) such as LSTM, it is not clear whether they keep the order of objects due to known issues of RNNs \cite{Pascanu2013}. In the field of neuroscience, it has been suggested that memorizing the order of objects that appear during navigation can identify the overall direction of the path, \textit{i.e.}, whether it is forward direction or return path \cite{Meilinger2016,Janzen2006,Wiener2012}. Therefore, by utilizing the order of objects, the robot may be able to make \rui{directionally consistent behavioral decisions.} 

Recent DRL navigation approaches for ObjNav have achieved remarkable improvements in performance using semantic and spatial knowledge about objects, such as object relationship, spatial object context, and object class hierarchy~\cite{Wei2018, Druon2020,Qiu2020,Du2021,Lv2020,Mayo2021}. However, the knowledge used in these studies is independent from the temporal context, which means that it lacks the information about the objects that are previously seen within the episode or it does not respect the order in which objects appeared.
%As a consequence, the agent cannot learn the relationship between temporal information and these semantic and spatial knowledge.
%As a consequence, the agent cannot acquire knowledge about these semantic and spatial information associated with the temporal information.
%This would result in the inability of the agent to project the qualitative topological knowledge of the environment during navigation.
%\blue{
%This would result in the inability of the agent to project the qualitative topological knowledge of the environment during navigation, or be an obstacle to the agent's understanding of the direction of its path.
%}
Although some of these studies have used Recurrent Neural Networks (RNNs), such as LSTM, it is not clear if they can understand the ordering of long-range  sequences~\cite{Pascanu2013}. %This would not result in to solve the problems described in the above.

Our goal is to investigate the effects of long-term observation of scenes and objects in ObjNav, and how we can effectively use them to improve navigation performance. To this end, we propose Object Memory Transformer (OMT) that consists of the following two key ideas, detailed in Fig.~\ref{figurelabel}. 
Firstly, we encode the current scene and object to fixed dimensional feature vectors, and store them in an external memory named Object-Scene Memory (OSM). This memory is capable of storing scene observations and object semantics throughout a long-term sequence in a temporally consistent manner.
Then, a Transformer network attends to the crucial scenes and objects, which are stored in OSM, and produces useful features for the RL agent.
% \blue{The key idea of our work is 
% \rui{%to learn the relationship between long-term memory and scenes/objects considered as a time series.
% to learn how to attend to crucial scenes/objects for navigation over long periods of observation
%to learn scenes/objects that are salient for navigation over a long period of observation.
%to project both scene observation and spatial objects context into a memory in a temporal manner.
% }}, which is detailed in Fig.\ \ref{figurelabel}. %Therefore, by utilizing the order of objects, the robot may be able to make \rui{directionally consistent behavioral decisions.} 
% To achieve this goal, we first encode long-term observations of scenes and objects \rui{then} store \rui{them} in an external memory named “Object-Scene Memory”, which is inspired by \cite{Fang2019}. This memory is capable of storing scene observations and object semantics throughout a long-term sequence in a temporally consistent manner.
We test our model on publicly available synthetic indoor environments, and show that OMT is able to navigate by taking efficient routes toward the target object. We also investigate whether our memory architecture improves the performance and efficiency in the ObjNav task.

%% file: related_works.tex
\section{Related Work}
%\subsection{Target-driven Navigation with Object Information }
\subsection{Object-goal Navigation}
%Object goal navigation (ObjNav) is one approach to goal-directed navigation, which is a variant of target-driven navigation~\cite{Zhu2017}. In ObjNav, an agent learns how to make decisions to reach the target object, such as move forward and turn right, based on a given word that specifies the target object and an observation image obtained at each time step. 

Yang et al.~\cite{Wei2018} proposed the first navigation RL agent to tackle ObjNav, using Graph Convolution Networks (GCNs)~\cite{Kipf2017} for incorporating scene priors. Since then, various studies have attempted to use semantic knowledge about objects to improve generalization performance to unseen targets and environments. In particular, the approaches that combine semantics and spatial knowledge about objects have achieved superior performance~\cite{Druon2020,Qiu2020,Du2021,Lv2020,Mayo2021}. However, in ObjNav a long-term sequence of observations has been rarely considered or exploited. In this paper, we propose a new external memory called Object-Scene Memory (OSM) to store the long-term time series of observed objects, and then use Transformer~\cite{Vaswani2017} to extract useful features for ObjNav.

\subsection{Memory Based Network and Transformer}
Recurrent Neural Networks (RNNs)~\cite{Jozefowicz2015} have been commonly used to capture time-series information. However, RNNs are known to be insufficient to handle long-term time-dependency. Therefore, a mechanism to store variables in an external memory outside the network has attracted attentions~\cite{Oh2016,Graves2014,Pritzel2017,Graves2016}. In the field of natural language processing (NLP), a method called Transformer~\cite{Vaswani2017} has been proposed and has shown its ability to process long sentences with many words based on the self-attention mechanism. The transformer architecture has become the de facto standard in NLP and has recently been applied to computer vision problems ~\cite{khan2021transformers,carion2020endtoend}.

\subsection{Navigation agents with memory}

Savinov et al.~\cite{Savinov2018} proposed a memory architecture called Topological Memory, referring to landmark-based navigation in animals. Khan et al.~\cite{khan2018memory} used the Differentiable Neural Computer (DNC)~\cite{Graves2016} in grid-world navigation. Zhu et al.~\cite{Pritzel2017} extended Neural Episodic Control (NEC) and proposed Episodic Reinforcement Learning with Associative Memory (ERLAM)~\cite{Zhu2020Episodic} to improve sample efficiency. ERLAM associates related experience trajectories by considering the relationship between states. Fang et al.~\cite{Fang2019} proposed Scene Memory Transformer (SMT) that stores long-term scene observations for none goal-directed navigation tasks. Du et al.~\cite{Du2020} proposed a Tentative Policy that avoids deadlock and loop by utilizing attention mechanisms for attending to all past actions stored in memory. Unlike these works, our Object-Scene Memory stores long-term object semantics that can support the agent to take an efficient trajectory to reach the target object in ObjNav.% In visual-language navigation, Transformer has shown its effectiveness in visual-semantics matching and handling long-term history~\cite{Fang2019,Wang2021,Pashevich2021}. However, again it has not been explored yet in ObjNav.

%% file: proposed_method.tex
\setcounter{figure}{1}
\begin{figure*}[t]
  \centering
  \includegraphics[width=0.90\textwidth]{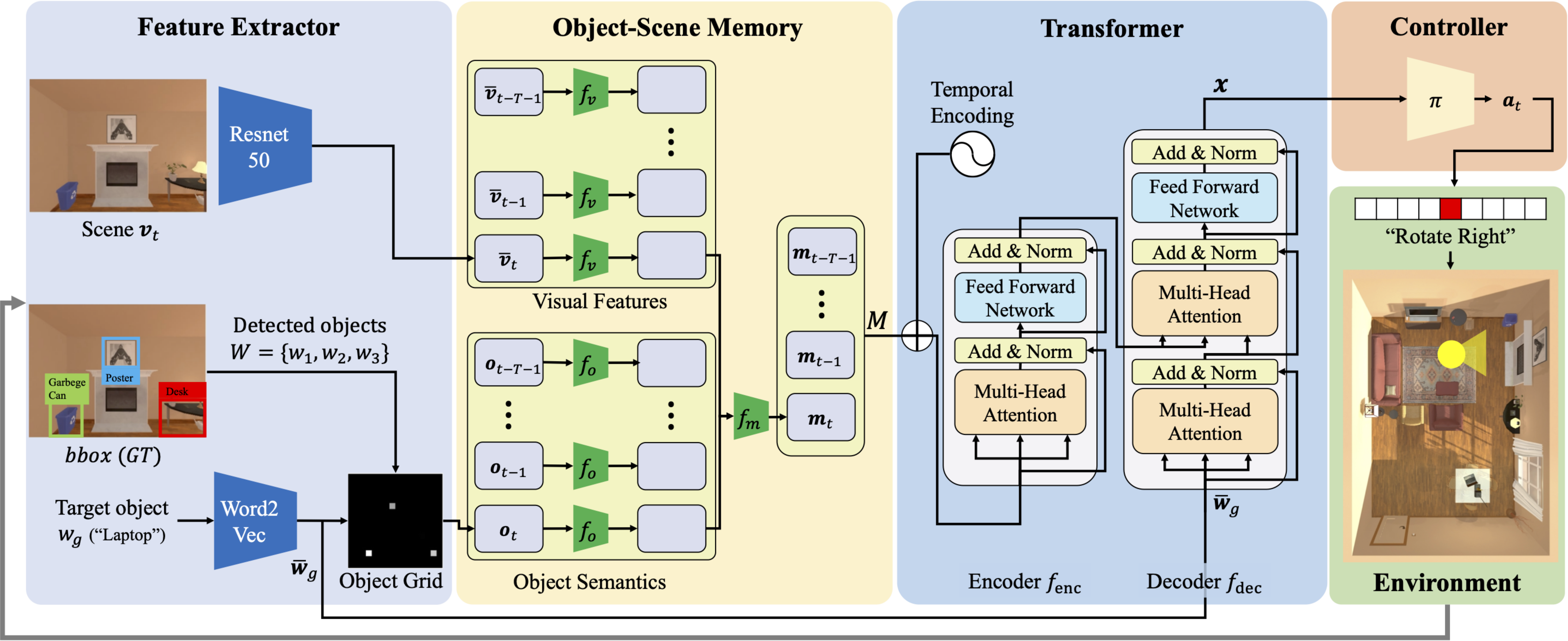}
  \caption{Architecture of OMT that consists of four steps: 
  1) Feature Extractor extracts visual features $\bar{\bm{v}}_t$ from RGB image $\bm{v}_t$ 
  %taken by the built-in camera on the robot 
  using ResNet-50. It also encodes the relationship between target object ${w}_g$ and the  objects observed in the scene,  %$W=\{\bm{w}_1,...,\bm{w}_{N^\mathrm{obj}}\}$ 
  $W=\{{w}_1,...,{w}_{N^\mathrm{obj}}\}$. This is done by computing the word embeddings of those objects using word2vec and cosine similarities between them, which will be assigned to the values in the object context grid, $\bm{o}_t$.
  2) OSM stores scene appearance $\bar{\bm{v}}_t$ and object semantics $\bm{o}_t$ for the last $T$-history length, and fuses both of those feature vectors into $\bm{m}_t$ using three learnable networks $f_{v},f_{o},$ and $f_m$. We apply this operation to all data stored in OSM, and obtains a set of features $M_{t}=\{\bm{m}_{t-T-1},...,\bm{m}_t\}$;
%   Scene memory processed by two 2D convolution layers with ReLU activation function $f_{v}$. Object memory and Scene memory are embedded in the end with embedding layer $f_{m}$ which contains a fully connected layer and ReLU activation function. 
  % The numbers in parentheses show the output dimensions.
  3) Transformer then takes $M_{t}^\mathrm{TE}$ so that the model can learn useful temporal  feature representations for ObjNav, which are conditioned by the target object;
  4) Finally, an A3C-based controller outputs action $\bm{a}_t$ to move in the 3D indoor environment.
  }
  \vspace{-3mm}
  \label{model}
\end{figure*}

\section{Proposed Method}\label{sec:propose}
%\konote{Briefly describe what we will show in this section.}
In this section, we describe our approach to learning a navigation agent for the ObjNav task. We start with explaining the task settings, and then introduce our novel model architecture that consists of two key components: 1) Object Scene Memory (OSM) that stores a history of observed scenes and objects; 2) Object Memory Transformer (OMT) that takes the history of observations from OSM, and produces useful features that will be taken as input to the subsequent RL agent.

%\subsection{Preliminary}
\subsection{Task Setting}\label{sec:task_setting} ObjNav is a task that considers a setting where an agent is placed at a random initial location in a 3D indoor environment, and required to reach a target object defined in the target class set $G=\{g_{0},...,g_{M}\}$ while minimizing the number of steps. %\konote{Describe more details about the goal sets. Is it word?}
Each task $\tau \in \mathcal{T}$ is represented by $\tau =\{e_{i},p_{i},g_{i}\}$, where an environment $e_{i}$ is given from a set of rooms $E=\{e_{0},\dots,e_{N}\}$, a starting position $p_{i}$, and a target object $g_i \in G$. 
At each state, the agent takes a RGB image in a first-person view, which we call {\it scene}, and {\it objects} in the form of bounding boxes, which exist in the scene. We use the terms {\it scene} and {\it object} in this manner.
% and bounding boxes along with objects class as an input.
%The \rui{inputs} the agent receives at each timestep consists of three information: 1) first-person view in the form of RGB image, which we call scene, 2) the objects \rui{class with its location in the view}, and 3) \rui{a} target object \rui{class}. 
%\blue{It is noted that other useful information, such as topographic maps or depth information about the environment, is not available in ObjNav \cite{Wei2018}.}

The action space from which the agent takes an action at each time step consists of the following nine commands: $\mathcal{A}=\{$\textit{Move Forward, Move Backward, Move Right, Move Left, Rotate Right, Rotate Left, Look Up, Look Down, Done}$\}$. The agent moves on the discretized scene space divided by $0.5\,\mathrm{m}$, and the rotation and the vertical tilt angle are $45$ degrees and $30$ degrees, respectively.

Each episode is considered to be success if the agent chooses {\it Done} action at positions where the target object is \textit{visible}, otherwise it is considered to be failure. Here, \textit{visible} becomes true when the target object is within threshold distance, which we set to be $1.5\,\mathrm{m}$, from the agent.% following~\cite{Qiu2020,Du2021,Lv2020,Du2020}.  
%two of the following three requirements are met: 1) the target object is in the camera's field of view at least partially, 2) the target object is not occluded by other objects, and 3) the target object is within threshold distance, which we set to be $1.5\,\mathrm{m}$, from the agent following~\cite{Qiu2020,Du2021,Lv2020,Du2020}. 

\subsection{Model Architecture} The proposed model architecture is shown in Fig.~\ref{model}. The agent model is based on Transformer for handling the long-term history of object semantics, which is inspired by Druon et al.~\cite{Druon2020} and Fang et al.~\cite{Fang2019}. %\rui{The principal idea throughout our model is that using the target features for both encoding objects semantics and a criterion for decision making.}
%Whereas, the principal idea throughout our model is that the target features are used for both encoding objects semantics and a criterion for decision making. This leads to object oriented decision making in ObjNav task.}
%which is accomplished by the Transformer.}
% and this purposes are described in 1) and 3).}
%The key idea of this model is that, (1. using w_g for encoding object semantics i memory and ....}
%Overall, the main idea of this model is that w_g is used to encode semantic information of objects and used as a criterion for decision making.
%The principal idea throughout the model is that the semantic information of the objects in the w_g criterion was encoded and used as a criterion for decision making.}
%It consists of five components. 1) Visual-Word Feature Extractor that extracts the features of the current scene $\bm{o_{t}}$ and target object $\bm{w}$. 2) Object Scene Memory that stores the long-term observations and objects semantics $M$ \rui{3) Temporal Encoding} 4) Transformer, and 5) \rui{RL}, which outputs an action $a_{t}$.
% This policy function is expressed as follows: 
% \begin{eqnarray}
% \label{eq:policy}
% a_{t} &\sim& \pi(a_{t} \mid \bm{o}_{t},\bm{w},M_{t-1}, \bm{\theta})
% \end{eqnarray}
Our method consists of 1) Feature Extractor, 2) Object-Scene Memory, 3) Transformer, and 4) Controller. We describe the above components in details.

%\textit{1) Visual-Word Feature Extractor}, \textit{2) Object Scene Memory}, \textit{3) Transformer} and \textit{4) Actor-Critic Layer}.

\subsubsection{\textbf{Feature Extractor}}
The feature extractor takes the current scene $\bm{v}_t$ and object words %$W=\{\bm{w}_1,...,\bm{w}_{N^\mathrm{obj}}\}$
$W=\{{w}_1,...,{w}_{N^\mathrm{obj}}\}$
and ${w}_g$ that specify the objects shown in the current scene and the target object, and outputs the features of the current scene and the target object.
%It takes a RGB image as its input and a class that specifies the target object, and extracts futures from them.
% The Visual-Word Feature Extractor takes as input $400\times300$ RGB image taken by the first-person camera and a word that specifies the goal. It then produces the visual features, word features, and ground-truth bounding boxes to describe the current scene.
To extract the visual feature, we use pre-trained ResNet-50~\cite{He2016} that takes the $400\times300$ RGB image $\bm{v}_t$ and produces $2048$-dimensional visual feature $\bar{\bm{v}}_t\in \mathbb{R}^{2048}$.
The semantic feature of the target, which is specified by the word ${w}_{g}$, is converted into $300$-dimensional word embedding vector $\bar{\bm{w}}_{g}\in \mathbb{R}^{300}$ using word2vec~\cite{Mikolov2013}. We do not update the parameters of these feature extractors during the whole process.
% The visual input to the network is a RGB $400\times300$ image. The input RGB image is then converted into 2048-d visual features using ResNet-50 pre-trained on the ImageNet dataset \cite{He2016}. On the other hand, the semantic features of the target and objects are given as a 300-d word embedding vector obtained by word2vec using Spacy toolkit~\cite{}. We frozen the parameters of these feature extraction networks during training.%\cite{Honnibal2017} 

In order to efficiently extract objects' semantics in the current view, we use the object context grid representation (object grid) proposed in~\cite{Druon2020}. The object grid $\bm{o}_t$ consists of an array with $16\times16$ cells, $\bm{o}_t\in \mathbb{R}^{16\times 16}$. We assign a value to each cell that indicates the similarities  between the target object and the detected objects. The positions to which these values are assigned correspond to the centers of the bounding boxes of the detected objects. The similarities between the target object and the detected objects are calculated by the cosine similarity as:
\begin{equation}
\label{eq:cos sim}
    \bm{o}_{i,j} =\frac{\bar{\bm{w}}_{i,j} \cdot \bar{\bm{w}}_{g}}{||\bar{\bm{w}}_{i,j}||  ||\bar{\bm{w}}_{g}||}, \nonumber
\end{equation}
where $\bar{\bm{w}}_{g}$ and $\bar{\bm{w}}_{i,j}$ denote the word embedding vectors of the target and an object whose center of bounding box detected at a location corresponding to $i$-th row and $j$-th column of the object grid $\bm{o}_t$, respectively. When no object located at $\bm{o}_{i,j}$, the value becomes to be $0$. 
As for the object set $W$, we assume the ground truth object labels and bounding boxes for all the objects in the view are available following~\cite{Druon2020,Qiu2020}.
%Note that using the \rui{context grid} it is faster to calculate visual semantic representation than using learning approaches such as \rui{Graph Convolution Network (GCN)} \cite{Wei2018} or visual attentions \cite{Mayo2021}, which may require additional effort and not that is not our focus in the paper.

\subsubsection{\textbf{Object-Scene Memory}}
Next, we introduce Object-Scene Memory (OSM) that has two functionalities: 1) storage of the visual appearance of the scene $\bar{\bm{v}}_t$ and the object semantics $\bm{o}_t$; 2) extraction of useful features from the stored data.
Focusing on the first one, OSM can be considered as a ring-buffer that can store $T$-history length. We prepare this ring-buffer for scene appearance $\bar{\bm{v}}_t$ and the object semantics $\bm{o}_t$ (see Fig.~\ref{model}).
Secondly, OSM fuses the spatial and object semantics to the 300-dimensional fused vector $\bm{m}_t\in\mathbb{R}^{300}$.
To that end, OSM employs three feature extractors $f_v,f_o,f_m$ that take scene appearance $\bm{v}$ and object semantics $\bm{o}$ as input. They fuse $\bm{v}$ and $\bm{o}$ to produce fused features $\bm{m}$ as:
\begin{equation}
    \bm{m}_t = f_m(f_{v}(\bar{\bm{v}}_t), f_o(\bm{o}_t)).
\end{equation}
% The both features for $u$-history length are stored in the OSM separately. % \rui{The memory length can be changed as desired.
%, but for the purposes of this explanation, we have chosen $32$ histories. 
We do this operation for all time frames of OSM, producing a set of fused features, $M_{t}=\{\bm{m}_{t-T-1},...,\bm{m}_t\}$.

\subsubsection{\textbf{Transformer}}
Transformer receives features stored in memory and produces useful features to the subsequent RL agent. We implemented temporal encoding and provided it to Transformer so that it can get a reference to temporal ordering of the memory. 

\textbf{Temporal Encoding. } Since we hypothesize that temporal information can improve the performance in the ObjNav task, we make use of the order of what the agent has observed in an episode by utilizing the temporal encoding. %Subsequently, Transformer network receive a memory with temporal information encoded.
%reference to the temporal dependencies in the memory.
% We implemented temporal encoding to make use of the order of the sequence, so that Transformer gets a reference to temporal dependencies in the memory by itself. 
% This temporal encoding is implemented right before the Transformer network. 
Among different implementations are proposed, we specifically use the trigonometric functions proposed in~\cite{Vaswani2017}. 
 Let $i$ be the current dimension of temporal encoding. Also, let $T$ be the size of $M$. Then, the memory with temporal order encoded, $M^{TE}$, can be formulated as:
% For this, we use the method using trigonometric functions proposed in~\cite{Vaswani2017}. The equation of this encoding is shown below.
\begin{eqnarray}
\label{eq:posienco}
% \bar{\bm{m}}_{t^\prime} &=& \mathrm{concat}(\bm{m}_t^\mathrm{sin}, \bm{m}_t^\mathrm{cos}) \\
% \bm{m}_t^\mathrm{sin} & = & \sin\left(\frac{\bm{m}_t}{10000^{2i/T}}\right)\\
% \bm{m}_t^\mathrm{cos} & = & \cos\left(\frac{\bm{m}_t}{10000^{2i/T}} \right) \\
%{\mathit{TE}_{(\mathit{pos},2i)}} & = & \sin\left(\frac{pos}{10000^{2i/T}}\right)\\
%{\mathit{TE}_{(\mathit{pos},2i+1)}} & = & \cos\left( \frac{pos}{10000^{2i/T}} \right),\\
%{\mathit{M}_{(\mathit{pos},2i)}}^{TE} & = & {\mathit{M}_{(\mathit{pos},2i)}} + \sin\left(\frac{pos}{10000^{2i/T}}\right)\\
%{\mathit{M}_{(\mathit{pos},2i+1)}}^{TE} & = & {\mathit{M}_{(\mathit{pos},2i+1)}} + \cos\left( \frac{pos}{10000^{2i/T}} \right),\\
{\mathit{M}_{(\mathit{pos},2i)}^{TE}} & = & \bm{m}^{pos}_{t(2i)} + \sin\left(\frac{pos}{10000^{2i/T}}\right)\\
{\mathit{M}_{(\mathit{pos},2i+1)}^{TE}} & = & \bm{m}^{pos}_{t(2i+1)} + \cos\left( \frac{pos}{10000^{2i/T}} \right),
\end{eqnarray}
where $pos \in{\{0, ...,T-1\}}$ and $\bm{m}^{pos}_{t}$ represent the location of the memory slot and feature $\bm{m}_{t}$ located at the $pos$-th slot of memory, respectively.

\textbf{Encoder-Decoder networks.} The temporally encoded scene and object context features are then fed into the Transformer encoder-decoder network~\cite{Vaswani2017} to obtain a useful representation for the RL agent.
% Here, we would like agent extract certain information from the interaction between spatiotemporal and landmark information according to target objects. Therefor we implemented a Transformer which described in detail in \cite{Vaswani2017}. 
Transformer employs a technique called self-attention to determine the relevance of each element to other elements in a temporal data and outputs contextualized features.

%Hence, it allows the agent to handle the temporal information without RNNs~\cite{Han2020a}, which are also designed to handle sequential input data and popularly used in ObjNav task.

The output of OSM along with temporal encoding $M^{\mathit{TE}}$ is taken as an input to the Tranformer encoder $f_\mathrm{enc}$, and the encoder extracts spatiotemporal and object semantics in the memory.
Then, the decoder $f_\mathrm{dec}$ produces useful representations $\bm{x}$ acquired from the attention between the encoded memory and the target object to the controller. %Using the target object as an input of decoder, it helps to promote generalizability to unknown environments rather than using current observation~\cite{Fang2019}.
%On the other hand, the decoder ($h$) takes the output of encoder and a target $g$ converted to word2vec ($\bm{w}$). 
%In order to promote generalizability, we also input the word embedding of the target object to the decoder.
% Using word2vec as a input of the decoder makes our model has more generalization to unknown scene than using observation as in \cite{Fang2019}.
% A simple formula is shown below.
\begin{eqnarray}
\label{eq:Transformer}
    \bm{x} &=& f_\mathrm{dec}(\bm{w}_g,f_\mathrm{enc}(M^{TE}))
    %Enco(M) &=& AttenBlock(M,M)\\
    %Deco(w,M) &=& AttenBlock(w,Enco(M))
\end{eqnarray}
%where $Enco$ and $Deco$ contains Attention block.

Note that the encoder takes masked attention for both self-attention and source-target attention when the memory slot is empty, i.e., when the current time step is shorter than the size of the OSM.

\subsubsection{\textbf{Controller}}
Finally, the features computed by the Transformer are passed to the controller. We specifically use Asynchronous Advantage Actor-Critic (A3C)~\cite{Mnih2016} for an RL agent. In our implementation, the model has two fully-connected layers and outputs both Q-values and the $| \mathcal{A} |$-length vector that has probability of each categorical action.
% contains two fully connected layers. %In this work, navigation policy learning is not the main focus, so we used the standard A3C \cite{Mnih2016} for training. 
% It outputs two values: the policy represented in the size of action set $\mathcal{A}$ and the Q-value for the current state. 

Overall, the policy at time $t$ can be expressed as:
\begin{eqnarray}
\label{eq:policy}
s_{t} &=& (\bm{v}_{t},\bm{w_{g}},M_{t-1}) \\
a_{t} &\sim& \pi(a_{t} \mid s_{t}, \bm{\theta}),
\end{eqnarray}
where $s_{t}$ and $a_{t}$ are the state and action at time $t$ and $\bm{\theta}$ are the model parameters.

%\begin{figure}[t]
%  \centering
%  \includegraphics[scale=0.35]{imgs/objscene.png}
%  \caption{Object and Scene Memory (OSM) architecture. OSM stores scene and objects semantics for the last $u$-history length. 
  %but theoretically an infinite number of time steps can be stored. 
%  Object memory processed by a fully connected layer with ReLU activation function $\varphi$. Scene memory processed by two 2D convolution layers with ReLU activation function $\psi$. Object memory and Scene memory are embedded in the end with embedding layer $\phi$ which contains a fully connected layer and ReLU activation function. 
%  The numbers in parentheses show the output dimensions. 
%  }
%  \label{obj and scene memory}
%\end{figure}

\subsection{Reward Design} 
The RL agent is trained to maximize the expected return defined as the expected cumulative reward $\mathbb{E_\pi}[\sum_{t=0}^{T}\gamma^t r_{t}]$. We define the reward function in Eq.~\eqref{eq:reward func}, which is a mixture of the one proposed in~\cite{Zhu2017} and~\cite{Ye2018} (details are described in~\cite{Druon2020}). 
% In brief, there are three types of rewards: a penalty for each time step so that the shortest path can be learned, an immediate reward calculated from the object similarity when the action results in approaching an target object, and a reward for successful navigation. The equation is shown in Eq.~\eqref{eq:reward func}.
% The reward function we use is as follows:
\begin{equation}
\label{eq:reward func}
    r = \begin{cases}
    5.0    &   \text{if success}\\ 
    S_{bbox}    &   \text{if $S_{bbox}$ is the highest in the episode}\\ 
    -0.01   &   \text{otherwise},
    \end{cases}
\end{equation}
where $S_{bbox}$ is computed by the ratio of the bounding box area in the field of view, and $S_{bbox}=1$ when the entire field of view is covered by the target object\footnote{If $S_{bbox}$ is not used, the agent immediately selects the ``$Done$" signal to avoid a decrease in the cumulative reward due to the huge state space and reward sparsity, resulting in a failure of learning. To avoid this, pre-training by imitation learning~\cite{Du2021,Lv2020,Du2020} or meta learning~\cite{Mayo2021,Wortsman2018} could be used.}.

The first condition encourages the agent to accomplish the task, and the second condition encourages the agent to move closer to the target object.
The third condition promotes to learn to navigate in the shortest path as the agent receives a penalty at each time step and tries to minimize it during training.

%\begin{figure*}[t]
%  \centering
%  \includegraphics[scale=0.50]{imgs/attention.png}
%  \caption{Visualization of the attention weight of the decoder when the "\textit{Done}" signal is emitted. The task is to find a book on a bed. The more white, the higher attention weight.}
%  \label{attention}
%\end{figure*}

%% file: experiments.tex
\section{Experimental Settings}
This section briefly describes the experimental settings including the dataset, evaluation metrics, and baseline models we used in the experiments to validate the proposed method. %For more information about implementation details, please refer to the longer pre-print version on the arXiv.  

\subsection{Dataset} We use the AI2-THOR framework (The House Of inteRactions)~\cite{Kolve2017}, which consists of $120$ different photo-realistic 3D indoor environments for our experiments. We use $80$ environments for training, i.e., $20$ environments from each of $4$ categories: Kitchen, Living room, Bedroom, Bathroom. The target class $G$ for each category is manually selected in advance from predefined $146$ objects class as follows: Kitchen: \{\textit{Toaster, Microwave, Fridge, Coffee Machine, Garbage Can, Bowl}\}; Living room: \{\textit{Pillow, Laptop, Television, Garbage Can, Bowl}\}; Bedroom: \{\textit{House Plant, Lamp, Book, Alarm Clock}\}; Bathroom: \{\textit{Sink, Toilet Paper, Soap Bottle, Light Switch}\}. Objects are placed in the room at random locations determined by the AI2-THOR framework. 

After training, we test the model in novel environments which are not included in the training environments to evaluate its generalization capability. Specifically, we used $20$ different environments, i.e., $5$ rooms from each of $4$ categories. Note that when the AI2-THOR environments were created, some targets were either not exist in the environment or not visible from any possible locations. In such cases, they were omitted from the target set for that environment.

\subsection{Evaluation Metrics}
We employ success rate (SR) and success weighted by path length (SPL)~\cite{Anderson2018} for our evaluation metrics. % to measure our model's performance. 
SR is defined as $\frac{1}{N}\sum_{i=1}^{N}\mathbb{I}_{\rm goal}^{i}$, and SPL is defined as $\frac{1}{N}\sum_{i=1}^{N}\mathbb{I}_{\rm goal}^{i}\frac{d^\ast}{d_{i}}$, where $N$ represents the total number of evaluation episodes and $\mathbb{I}_{\rm goal}^i$ is a success indicator, which is $1$ if the $i$-th episode is successfully finished, and $0$ otherwise. The success condition is described in Sec.~\ref{sec:task_setting}. The symbols $d^\ast$ and $d_{i}$ stand for the shortest path length in a given scene from the initial position to the target position, and the path length of $i$-th episode, respectively. In short, we can say a policy is {\it good} if it achieves high score on both SR and SPL.

\subsection{Models} We used the following models to compare the performance.
\subsubsection{\textbf{Random}} 
This model was implemented so that it can be shown how complex the experimental environments setup is.
% This is the simplest model for navigation. 
The agent uniformly samples one action from the set of actions $\mathcal{A}$ at each step. 
\subsubsection{\textbf{Scene Prior (SP)}}
This is an implementation of Scene Prior (SP)~\cite{Wei2018}, which is the simplest method of implementing semantic information of objects, implemented by~\cite{Druon2020}. SP uses Graph Convolution Networks (GCNs) to extract object semantics, where each node represents object class and edge represents relationship such as ``next'' or ``under/on.'' 
Our model differs from SP in dealing with observations: SP does not use spatiotemporal context of objects while ours does using OSM and OMT. SP takes four staked frames of scenes as input, while GCNs take only the current observation.
% However, SP do not contain object's spatiotemporal context while ours does. Note, this model takes four staked frames of observations as an input but GCNs is calculated based on current observation.
%This is an implementation of the model of \cite{Wei2018}, which is the simplest method of implementing semantic information of objects by GCN, which lean from FastText Database [Joulin et al 2016]. Note, this model takes four staked frames of observations as an input but GCN is calculated based on current observation.
\subsubsection{\textbf{Baseline}}
We use the model proposed in~\cite{Druon2020} for the baseline, as this method also employs the object context grid calculated with the ground truth object detector. While this method also uses semantic and spatial knowledge about objects as input, it does not explicitly utilize temporal information. Therefore, we can evaluate the effectiveness of our memory structure by comparing the results against this method. Note that this model takes four frames of observations as input. We also test against the model with three LSTM layers, which enables the model to handle temporal information. The LSTM layers take a single frame of observation as input, and the object grid is fused with it.
\subsubsection{\textbf{OMT}} We evaluate three models for our proposed method, whose size of the memory are $\{4, 16, 32\}$, respectively, to see the relationship between history length and its performance. We report the result with one layer of Transformer, but we have also obtained similar results with two layers. Please refer to Sec.~\ref{sec:propose} for more details.

%% file: results_discussion.tex
\section{Experimental Results}
In this section, we try to quantify the performance of OMT. Specifically, we try to answer the following questions: 
 \begin{itemize}
 	\item How does OMT perform on ObjNav in terms of SR and SPL compared against previous approaches?
 	%What is the performance and generalization achieved by long-term memory and Transformer?
    \item Which components of OMT contribute to the performance gain?
 \end{itemize}

\subsection{Performance of OMT}
\label{sec:performance}
\textbf{Comparison with baselines } First, we evaluated the performance of the proposed model. TABLE~\ref{tab: table1} shows the comparison between different approaches using the two different metrics. We can clearly see that our method outperforms prior works in both SR and SPL metrics.
The result of Random shows that the ObjNav task considers a huge navigation space and the agent cannot easily navigate to the goal state by chance. 
% The SP also has a poor performance of less than one-fifth of success rate. This is thought to be due to the lack of spatial context of the observed objects, which made it difficult for the agent to determine where to head.

Comparing the two Baseline models, Baseline showed SR and SPL of around $60\%$ and $20\%$, respectively. In addition, Baseline-LSTM $3$ layers, which can handle time-series information, showed a slight performance gain, i.e., $1.70\%$ higher SR and $3.13\%$ higher SPL than Baseline.

Focusing on our model, it achieves the best performance in terms of SR and SPL. Our model with $4$ histories, which has the same history length as Baseline, shows higher SR ($71.13\%$) than Baseline and Baseline-LSTM $3$ layers. Furthermore, when the history length is long, our agent model is more likely to take the shortest path. In fact, our model with $32$ histories shows the best SPL ($27.51\%$), which suggests that long-term histories can improve navigation efficiency by choosing a shorter path to the target. 

\begin{table}[t]
 \begin{center}
 \caption{Evaluation results on unknown scenes and known objects}
 %\vspace{-1.5mm}
 \label{tab: table1}
 
 \footnotesize
 \scalebox{1.0}{
	\begin{tabular}{lcc} \toprule
    Method & SR [$\%$]. & SPL [$\%$]\\ \midrule
    Random & 6.10 & 1.30 \\ 
    SP \cite{Wei2018} & 18.24 & 4.39 \\
    Baseline \cite{Druon2020} & 61.55 & 20.83\\
    Baseline - LSTM 3 layer \cite{Druon2020} & 63.25 & 23.96\\ \midrule\midrule
    OMT - 32 hist. & 69.39 \scriptsize{(0.16)} & \textbf{27.51} \scriptsize{(0.10)}\\  
    OMT - 16 hist. & 70.00 \scriptsize{(1.56)} & 27.12 \scriptsize{(0.22)}\\
    OMT - 4 hist. & \textbf{71.13} \scriptsize{(1.42)}& 26.66 \scriptsize{(0.10)}\\ \bottomrule
	
	\end{tabular}
}
\end{center}
\vspace{-1.5mm}
\end{table}

\textbf{Comparison with SOTA}
We also compared the performance against the current state-of-the-art (SOTA) techniques~\cite{Du2020,Qiu2020} based on the SR and SPL metrics reported in the papers. Note that the experimental settings used in those works are not exactly the same as ours. Thus some care must be taken when comparing each other.

Qui et al.~\cite{Qiu2020} proposes a navigation method that leverages a hierarchical object relationship in the ObjNav task. While the method is tested on the different grid step size ($0.25\,\mathrm{m}$) from ours ($0.5\,\mathrm{m}$), its action space is $| \mathcal{A} | =5$. Thus it explores smaller action space and makes the task easier. The reported scores are SR of $65.3\%$ and SPL of $21.1\%$. Hence, we can conclude that our method works better than \cite{Qiu2020}, as our method achieves better performance for both of the two metrics on a harder problem (action space is bigger).

Du et al.~\cite{Du2020} proposes the memory-augmented tentative policy networks to promote the agent to escape from deadlock states, such as looping or being stuck. Although we could not compare SPL as their definition is different from ours, the reported SR of $69.3\%$ is $2\%$ lower than our best result. Thus, we can conclude that our OMT outperforms this SOTA method as well.

\begin{table}[t]
 \caption{Results  without each component.}
 \vspace{-2.0mm}
 \label{tab: table2}
 \begin{center}
 \footnotesize
 \scalebox{0.9}{
	\begin{tabular}{lcc|cc} \toprule
	\multirow{2}{*}{\begin{tabular}{l}Method\end{tabular}} & \multicolumn{2}{c|}{32 hist.} & \multicolumn{2}{c}{4 hist}\\
	 & SR [$\%$] & SPL [$\%$] & SR [$\%$] & SPL [$\%$]\\ \midrule
    OMT(full) & 69.39 & \textbf{27.51} & 71.13 & 26.66\\ 
    w/o Object Memory  & 67.99 & 25.49 & \textbf{71.23} & 26.69\\
    w/o Scene Memory & 65.38 & 23.15 & 62.60 & 25.10\\
    w/o Scene & 62.30 & 24.28 &65.59 &24.81 \\ 
    w/o Temporal Encoding & 59.69 & 21.95 &61.41 &23.29 \\ 
    w/o Transformer & 64.27 & 25.43 &68.84 &25.71 \\ \bottomrule
    %Ours-dropout \Rui{(may not needed)} & 70.65 & 27.75\\ 
    %Ours-2 layer \Rui{(may not needed)} & \textbf{71.09} & \textbf{28.63}\\ \bottomrule
	\end{tabular}
 }
 \vspace{-5mm}
\end{center}
\end{table}

\subsection{Ablation study---Effects of each component} 
To analyze what component of our method contributes to the performance gain, we conducted an ablation study by removing one component from the full model and measuring the performance in the two metrics as in Sec.~\ref{sec:performance}. % to find out which of those were critical for navigation. 
%For example, Fig.~\ref{simplemodel} shows the {\it w/o Object Memory} model in our experiment. 
The results are shown in TABLE~\ref{tab: table2}.
The model {\it w/o Object Memory} removes the context grid, which encodes object semantics in the current scene, from OSM and directly passes to the Controller. Likewise, for a model {\it w/o Scene Memory}, the current scene is also concatenated. On the other hand, the {\it w/o Scene}, the current scene is not used, and only the context grid is used as input. In the model of {\it w/o Temporal encoding}, the temporal order is given by a soft manner like~\cite{Fang2019} with using an exponential function which is concatenated with visual features. In the {\it w/o Transformer} model, Transformer is replaced with two fully connected layers and ReLU function.% \rui{(fig.s in supplementary material).}
%This way, the reference of temporal ordering is provided more precisely and clearly to memory, not just differentiating between recent and old observation.%The reverse is also true for Scene Memory.

To evaluate how much scene information (first-person views) contributes to the navigation performance, we first see the results of {\it w/o Scene Memory} and {\it w/o Scene}. To our surprise, we can see that both methods perform surprisingly well even without scene information, whose scores are comparable with the baseline model with LSTM. This shows that a decent level of navigation is possible only using spatiotemporal knowledge of objects without recognizing a whole scene image.
Next, we removed the object memory from the full model and saw SR dropped $1.40\%$ for $32$ histories, but it was $0.1\%$ higher than $4$ histories. On the other hand, when all the components are incorporated into our model with $32$ histories, it shows the highest SPL score. This indicates that navigation efficiency is the consequence of interacts between scene, object, and memory. Especially, the temporal encoding of Transformer was found to have the most prominent impact on navigation performance, indicating that the temporal ordering of histories in memory is vital for ObjNav.

\subsection{Qualitative study}
To qualitatively show the effectiveness of our key idea in using the long-term history in ObjNav, we plotted agents' trajectories for the episodes where the start and goal are far from each other. In these cases, it is likely that there are obstacles in its route and that the target is not visible from the start location.    

\begin{figure}[t]
  \centering
  \includegraphics[width=0.44\textwidth]{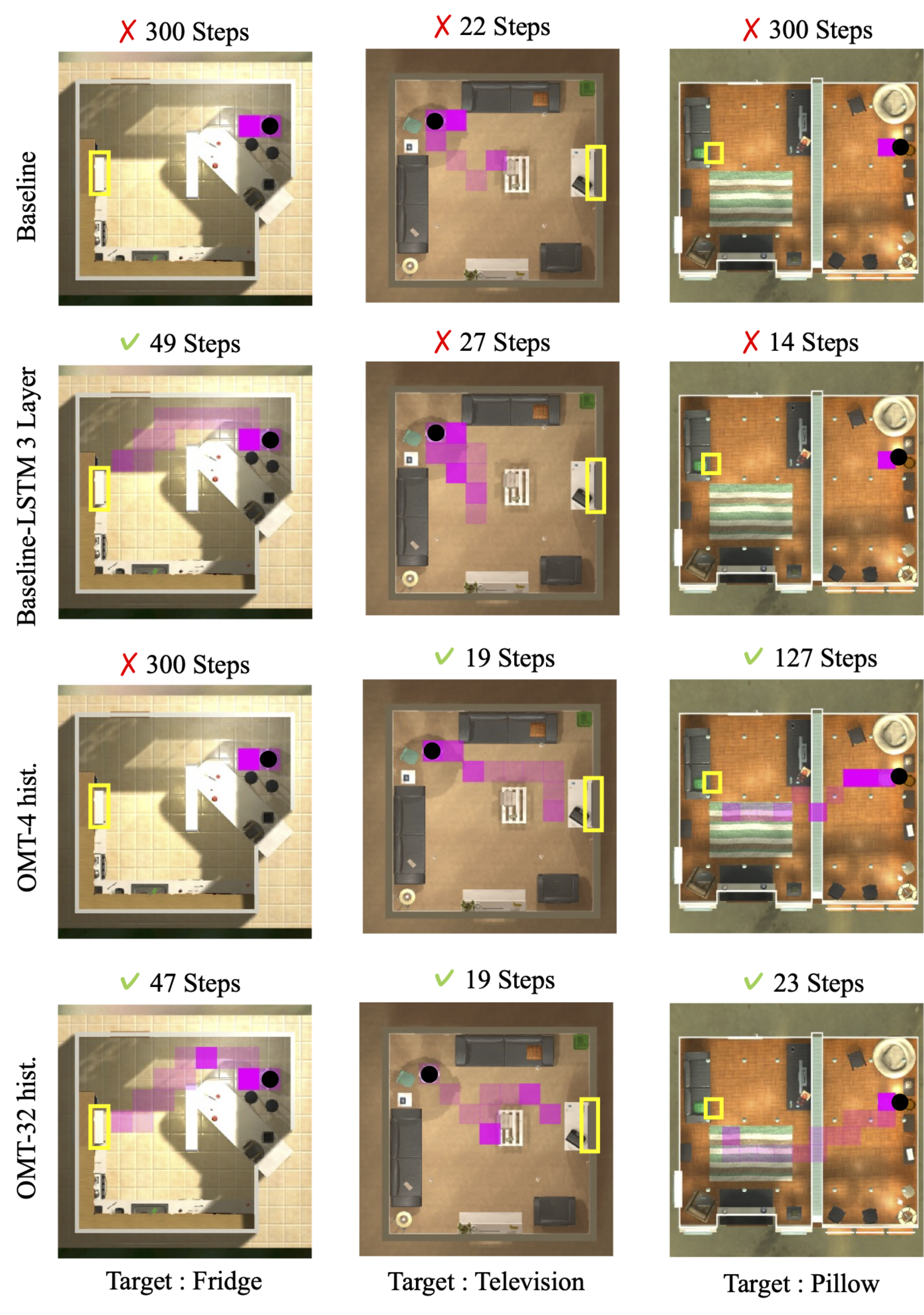}
  \vspace{-0.5mm}
  \caption{Qualitative results. The top-down map shows the trajectories of different methods.
  %Baseline, Baseline-LSTM $3$ Layer, OMT-$4$ hist., and OMT-$32$ hist.. 
  The initial location of the agent is shown as a black dot, and the target object is circled by a yellow box. Trajectories are indicated by purple squares of varying density. The thicker color, the more number of times the agent stays at the location. The number above each top-down map shows the total number of steps in this episode.}
  %(Top row) OMT-$32$ hist. achieved the {\it Fridge} even once got stuck in the obstacle, while Baseline and OMT-$4$ hist.~was unable to move. (Middle row) OMT-$4$ hist.~and OMT-$32$ hist.~reach the {\it Television}, while Baseline and Baseline-LSTM $3$ layer select {\it Done} signal from outside the threshold range, i.e., the distance estimation is wrong. (Bottom row) The target object is placed at unobservable location from the agent's start position because of the wall in the middle. While OMT-$32$ hist.~reaches the target, other methods struggle to escape the loop at the beginning of the episode.}
  \label{fig:qual_study}
  \vspace{-5mm}
\end{figure}

Fig.~\ref{fig:qual_study} shows the top-down maps with the trajectories of Baseline, Baseline-LSTM $3$ Layer, OMT-$4$ hist., and OMT-$32$ hist. Focusing on the {\it Fridge} results (left column), Baseline and OMT-$4$ histories move straight toward or move left to get close to the target from the start, but are unable to move due to an obstacle. In contrast, OMT-$32$ histories once got stuck in an obstacle, but is able to get out of it and reach the target. We believe this is because OMT can perceive the continuity and changes of observations over time. In particular, the OMT recognizes its deadlock state when the same observations continue to be perceived  over a long time. Similarly in {\it Television} (middle column), OMT-$32$ histories can recover to move to goal position even it gets stuck in bumping into the desk before the television, while Baseline and Baseline-LSTM $3$ layer select {\it Done} signal from outside the threshold range.
%, i.e., the distance estimation is wrong.
% This is thought to be due to the fact that it avoided deadlock by using long-term history. This tendency could be seen in (Fig.~\ref{compare} middle), where our-$32$history model bumped into a desk. 
In {\it Pillow} (right column), where the agent is unable to see the target from the start location because of the wall in the middle, OMT-$32$ histories once gets stuck in a loop at the beginning of the episode, but is able to get out of it and reached the target while Baseline stuck in a rotation loop until the episode exceeds the maximum number of steps.
These results suggest that our proposal of utilizing the long-term information can help guide the agent in complex ObjNav tasks, where the agent cannot see the goal object at the start position, especially in settings where the distance between the start and goal is long.

%% file: conclusion.tex
\vspace{-0.5mm}
\section{Conclusion}
\vspace{-0.5mm}
In this paper, we investigated how long-term histories of objects and the first-person views can improve navigation performance in ObjNav task. To this end, we presented Object Memory Transformer (OMT) that utilizes the histories of information that an agent observes during navigation. The proposed method consists of two key components: 1) Object-Scene Memory (OSM) that enables to store long-term scene and object semantics, and 2) Transformer that attends to salient objects in the sequence of previously observed scenes and objects.
Our experimental evaluation demonstrated that OMT can achieve state-of-the-art performance on the AI2-THOR benchmark compared against prior works that do not use long-term histories or use them in different manner. Additionally, we showed that just using the long-term histories does not help improve the performance, but combining them with temporal encoding is essential. 